\documentclass{./styles/svproc}
\usepackage{url}

\usepackage{xcolor} 
\usepackage{amsmath,amsfonts} 
\usepackage{float} 
\usepackage{graphicx}  
\usepackage{soul}
\usepackage{array}
\usepackage{bm}  

\usepackage[linesnumbered,ruled]{algorithm2e}

\SetKwInput{KwInput}{Input}                
\SetKwInput{KwOutput}{Output}              
\SetKwInput{kwConstraint}{Constraint}      
\SetKwBlock{KwTA}{TA}{end}
\SetKwBlock{KwDH}{Data Holders}{end}
\SetKwBlock{KwCSP}{CSP}{end}
\SetKwFunction{FFedPCATrain}{FedPCA.train}
\SetKwFunction{FFedPCAPredict}{FedPCA.predict}
\SetKwFunction{FFedMPCAPredict}{FedMPCA.predict}
\SetKwProg{Fn}{Function}{:}{End}

\usepackage[colorlinks=true,linkcolor=blue,citecolor=blue,allcolors=blue]{hyperref}

\begin{document}
\mainmatter

\title{Towards federated multivariate statistical process control (FedMSPC)}
\titlerunning{FedMSPC}
\author{Du Nguyen Duy\and David Gabauer \and Ramin Nikzad-Langerodi$^*$}
\authorrunning{Du et al.}
\tocauthor{Nguyen Duy Du, David Gabauer, and Ramin Nikzad-Langerodi}
\institute{Software Competence Center Hagenberg, Austria\\
\email{\{du.nguyen.duy, david.gabauer, ramin.nikzad-langerodi\}@scch.at}
}
\maketitle

\begin{abstract}
The ongoing transition from a linear (produce-use-dispose) to a circular economy poses significant challenges to current state-of-the-art information and communication technologies. In particular, the derivation of integrated, high-level views on material, process, and product streams from (real-time) data produced along value chains is challenging for several reasons. Most importantly, sufficiently rich data is often available yet not shared across company borders because of privacy concerns which make it impossible to build integrated process models that capture the interrelations between input materials, process parameters, and key performance indicators along value chains. In the current contribution, we propose a privacy-preserving, federated multivariate statistical process control (FedMSPC) framework based on Federated Principal Component Analysis (PCA) and Secure Multiparty Computation to foster the incentive for closer collaboration of stakeholders along value chains. We tested our approach on two industrial benchmark data sets - SECOM and ST-AWFD.  Our empirical results demonstrate the superior fault detection capability of the proposed approach compared to standard, single-party (multiway) PCA. Furthermore, we showcase the possibility of our framework to provide privacy-preserving fault diagnosis to each data holder in the value chain to underpin the benefits of secure data sharing and federated process modeling.


\end{abstract}

\section{Introduction}

The importance of data exchange along value chains has been broadly recognized for mastering the transition from linear to circular economy \cite{pagoropoulos2017,pennekamp2019}. However, as individuals and corporations are increasingly concerned about how their data are being used, the emphasis on data privacy and security has become a major global topic. There are now data protection obligations that organizations must strictly follow \cite{EU2022}\cite{DLA2022}. Therefore, it is challenging, if not impossible, in many situations to transfer data across company borders. This landscape poses new challenges that traditional information and communication technologies in general, and process modeling approaches in particular, cannot handle appropriately.

More specifically, traditional process modeling workflows usually involve collecting and fusing data into a common site where a data-driven model is built. However, this is no longer feasible since data are not owned by a single entity but rather generated and distributed among different companies along a value chain. On the one hand, organizations do not want to share private data because of the fear of trade secrets leaks, and on the other hand due to regulations or geographical restrictions. As a result, even though sufficiently rich process data is available, they often exist in small and fragmented silos and cannot be integrated to enable a broader view of the whole value chain. This is a major obstacle in areas where it is well known that material properties, process parameters, and KPIs are intercorrelated across company borders, for example, in the steel or paper industry \cite{winning2017}\cite{molina2018}.

A functional solution to overcome the problem of data fragmentation and isolation is Federated Learning (FL), a concept first proposed by Google in 2016 \cite{konecny2016}. The main idea behind FL is to build a centralized model based on data scattered among multiple parties without requiring participants to share sensitive information. In the meantime, FL has gained increasing attention, both from research and industry perspectives. However, a preliminary literature review shows that most of the current work is focused primarily on deep neural networks and their application to computer vision problems \cite{li2020}\cite{qinbin2022}. 

Multivariate Statistical Process Control (MSPC) is an umbrella term for a set of advanced statistical methods for modeling, monitoring, and controlling the operating performance of processes that are widely adopted in the process industry. More specifically, MSPC techniques extract features from high-dimensional and highly correlated process data by means of latent variables (LVs) based modeling techniques. The models are then used to monitor processes in real-time, assess their performance, and identify deviations from normal operating conditions (NOC). Therefore, MSPC provides a basis for increasing process security, sustainability, and continuous improvement. Although some recent research efforts have been devoted to adopting federated learning in the field of MSPC, limited progress has been made \cite{hartebrodt2021}\cite{grammenos2020}. Previous studies have focused on federated principal component analysis (PCA) \cite{chai2021}. However, the application of PCA to MSPC, i.e. for online monitoring, fault detection, and diagnosis has not been proposed so far. In addition, multiway PCA (MPCA), an extension of PCA for modeling batch process data has not been investigated, and thus, to the best of our knowledge, the FL paradigm has so far not been adopted for building MSPC-type process models across company borders that preserve the privacy of each contributing party. Moreover, the incentive mechanism, an essential aspect of FL, has not been discussed in the existing literature. A fair value-distribution structure is critical to motivating the different parties to actively collaborate in the model training and inference process \cite{yang2019}. The collaboration might be wasteful without meaningful incentives because the participating parties will not carry out efficient contributions.

In this work, we propose a general federated multivariate statistical process modeling framework (FedMSPC) where different companies along a value chain can together build a shared process monitoring model in a federated and privacy-preserving manner. To fully protect confidential data, FedMSPC uses a combination of two privacy techniques: Differential Privacy (DP) \cite{ji2014} and Secure Multiparty Computation (SMC) \cite{bonawitz2016}. Each participant will preprocess and encrypt data in his local environment using a well-designed DP method. Then all encrypted data are transferred to a third-party server, which securely aggregates these data and trains an MPCA model following the batch-wise unfolding of the joint (encrypted) data matrix. Finally, using their private key, each participant will decrypt the federated output of the model to get the actual and relevant results. More specifically, each party will get the portion of the shared loadings matrix that corresponds to the variables that they contribute. This information is secretly known only by the party and is hidden from all other participants. In addition, all parties will share the explained variance corresponding to the selected principal components. Using such results, all data holders can collaboratively estimate the scores, process-monitoring statistics (e.g. Hotelling's $T^2$ and $Q$-statistic) as well as variable contributions, thereupon conduct fault detection and diagnosis.

In order to showcase the feasibility of the framework, we propose Federated Principal Component Analysis (FedPCA), which is based on the idea of Federated Singular Value Decomposition (FedSVD) proposed in \cite{chai2021}, as the modeling method. FedSVD basically provides lossless privacy guarantees and is thus ideally suited for building federated MSPC models. However, in \cite{chai2021}, the authors employ FedPCA under a horizontally partitioned scenario, where the data from the contributing parties share the same feature rather than the sample space (i.e. horizontal FL). However, in value chains, input materials are processed sequentially by different companies and the corresponding data is thus vertically partitioned, i.e. the data share the same sample space but different feature spaces. Therefore, we will concentrate on vertical FedPCA instead. In addition, we will investigate the application of FedPCA in processing batch data. 

\section{Methodology}

\subsection{Multivariate statistical process control}

Multivariate statistical process control (MSPC) is a widely applied approach for process monitoring. For high-dimensional and highly correlated process data, one of the most well-known and widely adopted MSPC methodologies is Principal Component Analysis (PCA). 

\subsubsection{Principal Component Analysis (PCA)} is often applied in MSPC to transform a dataset with highly correlated variables into an uncorrelated dataset while preserving only the systematic variation. There are various techniques for building a PCA model, SVD being the most popular one. Suppose the original dataset denoted as $\bm{X} \in \mathbb{R}^{m \times n}$ contains $m$ observations and $n$ process variables. SVD corresponds to the following decomposition:
\begin{equation}
\begin{split}
\bm{X} & = \bm{U} \bm{\Sigma} \bm{V}^{T} \\
  & = \begin{bmatrix} \bm{U}_r & \bm{U}_0 \end{bmatrix} \begin{bmatrix} \bm{\Sigma}_r & 0 \\ 0 & \bm{\Sigma}_0 \end{bmatrix} \begin{bmatrix} \bm{V}_r & \bm{V}_0 \end{bmatrix}^T,
\end{split}
\end{equation}
where $\bm{U} \in \mathbb{R}^{m \times m}$ is the left singular matrix, $\bm{\Sigma} \in \mathbb{R}^{m \times n}$ is the diagonal matrix whose diagonal elements are singular values, and $\bm{V} \in \mathbb{R}^{n \times n}$ is the right singular matrix. $\bm{V}_{r} \in \mathbb{R}^{n \times r}$ derived from $\bm{V}$ is called the loadings matrix. The number of principal components $r$ can be determined based on a certain criterion, e.g. the cumulative explained variance, and usually, it holds that $r \ll n$. The loadings are the coefficients of the variables from which the principal components are computed. The sign of the loading shows whether the correlation between the principal component and the variable is positive or negative while its absolute value indicates how strongly the variable influences the principal components. Therefore, they are often used to quantify variable importance.

Projecting $\bm{X}$ onto the subspace spanned by the selected principal components ($PCs$) reduces the dimensionality of the column space from $n$ to $r$. The result of the transformation is the scores matrix $\bm{T}_{r} \in \mathbb{R}^{m \times r}$ 
\begin{equation}
\bm{T}_{r} = \bm{X}\bm{V}_{r}.
\end{equation}
The reconstruction of $\bm{X}$ can be estimated from $\bm{T}_{r}$ and $\bm{V}_{r}$:
\begin{equation}
    \hat{\bm{X}} = \bm{T}_{r}\bm{V}_{r}^{T}
\end{equation}
The residual matrix $\bm{E}$ is defined as the errors between $\bm{X}$ and $\hat{\bm{X}}$ and can be calculated as:
\begin{equation}
    \bm{E} = \bm{X} - \hat{\bm{X}} = \bm{X} - \bm{X}\bm{V}_{r}\bm{V}_{r}^{T}
\end{equation}

\subsubsection{Multi-way principal component analysis (MPCA)} is an extension of PCA for monitoring batch processes that are broadly seen in industries where batch and semi-batch process operations are common, for example, in the chemical or pharmaceutical industry \cite{nomikos1994}\cite{nomikos1995multivariate}. 

Suppose each batch run has $J$ variables measured at $K$ time intervals. Similar data exist in the same form for each of the $I$ batch runs. Therefore, all the data can be arranged in an array $\bm{X} (I \times J \times K)$. 

\begin{figure}[H]
  \centering
  \includegraphics[width=1.0\textwidth]{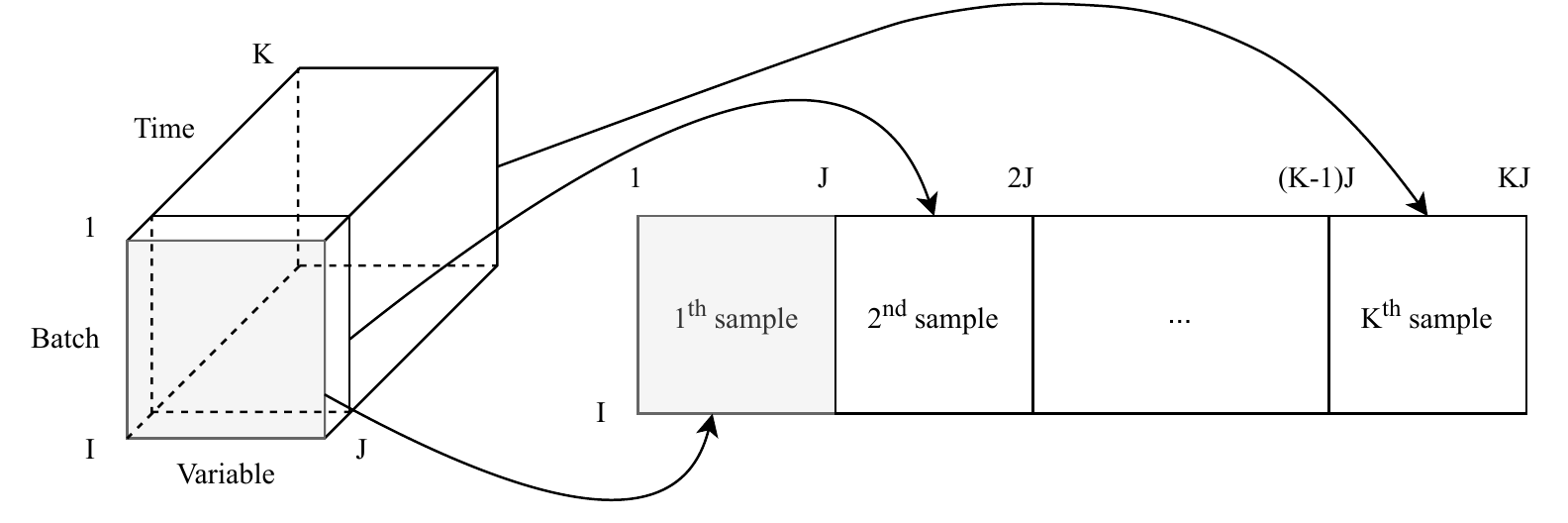}
  \caption{An illustration of the batch-wise unfolding method.}
  \label{fig:unfolding}
\end{figure}

In order to apply PCA, the dataset has to be converted into a two-dimensional array. There are multiple ways to unfold a 3D dataset. However, the most meaningful approaches are batch-wise and variable-wise unfolding. In the proposed approach, we will employ batch-wise unfolding since variable-wise unfolding is not feasible for federated MSPC. In batch-wise unfolding, the 2D array is formed by unfolding the array $\bm{X}$ so that each of its vertical slices contains the observed variables for all batches at a specific time instance. The result is a 2D matrix of the shape $(I \times KJ)$. An illustration of this unfolding technique is shown in Figure \ref{fig:unfolding}.

After the data is unfolded, PCA is performed to retrieve the scores and loadings matrices. $T^2$, $Q$-statistics, and contribution plots can then be calculated in a similar manner as for PCA.

From an operational point of view, it is preferable to monitor the batch as it progresses in order to anticipate process faults and to take timely actions to prevent out-of-batch-specification events. However, in this situation, a major obstacle is that the new batch is required to have $KJ$ columns such as the NOC data used for training the model. This is impossible when the batch has not been completed because at time interval $k < K$ the new batch $x$ only has $kJ$ columns. In \cite{nomikos1995multivariate}, the authors proposed a simple solution to overcome this problem which is to use only the portion of the loadings matrix that corresponds to the elapsed time period until the current time interval $k$ to calculate the new scores vector
\begin{equation}
    t_r^{[k]} = x\tilde{\bm{V}_r}(\tilde{\bm{V}_r}^{T}\tilde{\bm{V}_r})^{-1},
\end{equation}
where $\tilde{\bm{V}_r} = \bm{V}_r[1:kJ]$ contains the first $kJ$ columns of $\bm{V}_r$. In the proposed approach, we will use this same approach to handle the situation where batches are incomplete.

\subsubsection{Fault detection and diagnosis} One of the most popular applications of PCA-based MSPC is fault detection and diagnosis. This is often done based on control chart statistics, such as Hotelling's $T^2$ and the so-called $Q$-statistic. While a high $Q$-statistic indicates a change in covariance structure, a high $T^2$ indicates that although the sample is described well by the model (i.e. through a linear combination of the loadings) it is unusual in terms of the linear combination. Suppose there is a PCA model generated from NOC data. When a new sample $x \in \mathbb{R}^{1 \times n}$ arrives, $T^{2}$ and $Q$ can be calculated as:
\begin{equation}\label{eq:T2_calculation}
T^{2} = t_{r}\bm{\Lambda}^{-1}_{r}t_{r}^{T},
\end{equation}
where $t_{r} = x\bm{V}_{r}$ and $\bm{\Lambda}_{r} = \bm{\Sigma}_{r}^{2}$ and
\begin{equation}\label{eq:Q_calculation}
Q = \sum_{i=1}^{n} (x_i - \hat{x}_i)^2.
\end{equation}
The upper confidence limit for the $T^2$ statistic can be computed from the $F$-distribution.
\begin{align}
T^{2}_{\alpha} = \frac{r(m - 1)}{r-m}F_{r,m-r,\alpha}, 
\end{align}
where $m$ and $r$ denote the number of NOC samples and the number of principal components, respectively. The upper confidence limit for the $Q$-statistic can be computed from its approximate distribution.
\begin{align}
\begin{aligned}
Q_{\alpha} =& \theta_1 \left( 1- \frac{\theta_2 h_0 (1-h_0)}{\theta^{2}_{1}} + \frac{z_{\alpha} \sqrt{2 \theta_2 h^{2}_{0}}}{\theta_1} \right)^{1/h_0}\\
\theta_i =& \sum_{j = r + 1}^{l} (\bm{\Lambda}_{j,j})^i, i = 1, 2, 3\\
h_0 =& 1 - \frac{2\theta_1 \theta_3}{3\theta^2_2},
\end{aligned}
\end{align}
where $z_{\alpha}$ is the standard normal deviate corresponding to the upper $(1-\alpha)$ percentile, $\bm{\Lambda}_{jj}$ is the eigenvalue associated with the $j^{th}$ loading vector, and $l$ is the number of non-zero eigenvalues calculated from the data.

A sample might be considered faulty if either $T^{2}$ or $Q$ exceeds the predefined control limits $T^{2}_{\alpha}$ or $Q_{\alpha}$. When a fault is detected, contribution plots can be generated to show how input variables contribute to $T^2$ and $Q$ \cite{nomikos1995multivariate}. The contributions of the variables to $T^2$ and $Q$ statistics are calculated as
\begin{equation}
    T^{2}_{cont} = x\bm{V}_{r}\bm{\Sigma}^{-1}\bm{V}_{r}^{T} = t_{r}\bm{\Sigma}^{-1}\bm{V}_{r}^{T}
\end{equation}
and
\begin{equation}
    Q_{cont} = (x - \hat{x})^2.
\end{equation}
Variables with high contributions are diagnosed as candidates for the cause of the fault.

\subsection{Federated multivariate statistical process control (FedMSPC)}

\begin{figure}[t!]
  \centering
  \includegraphics[width=0.9\textwidth]{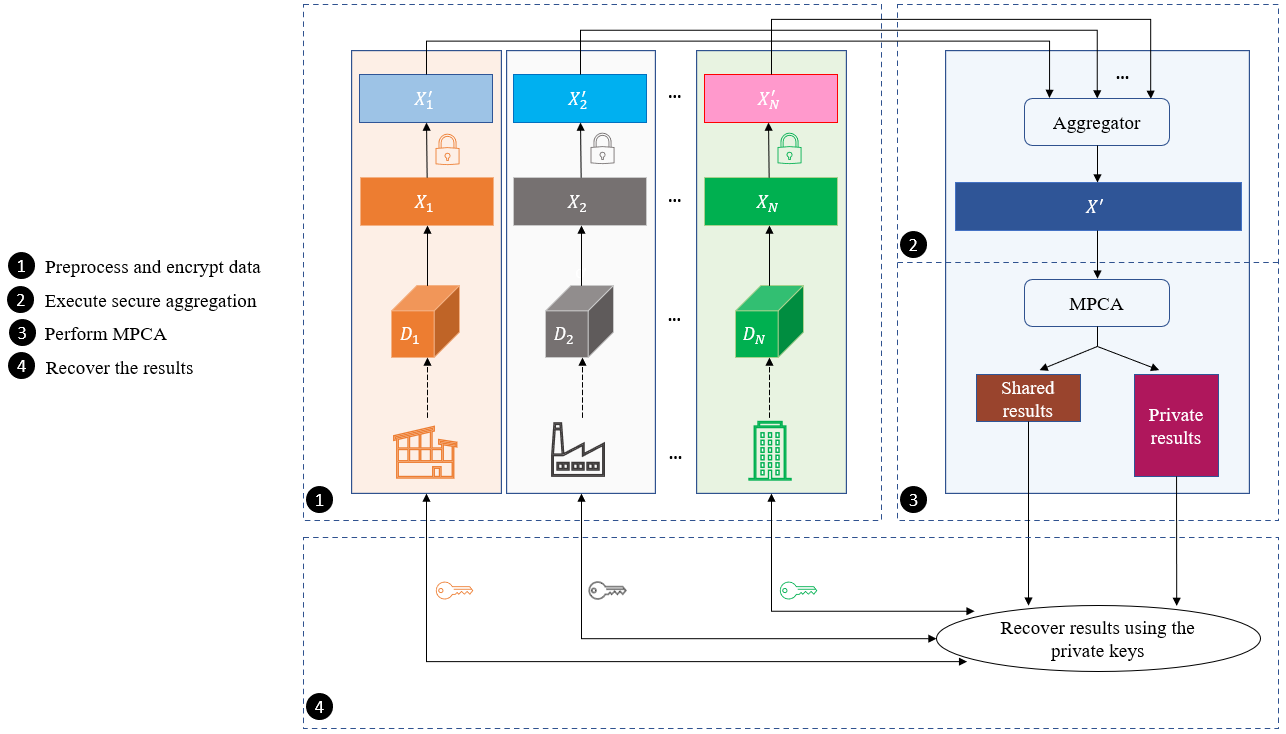}
  \caption{The architecture of the Federated Process Modeling framework.}
  \label{fig:arch}
\end{figure}

Figure \ref{fig:arch} illustrates the proposed framework. Assume we have $g$ data holders with the $i$-th data holder owning data matrix $\bm{X}_i \in \mathbb{R}^{m \times n_i}$ and these data holders aim at fitting a PCA model on the concatenated matrix $\bm{X} = [\bm{X}_1, \bm{X}_2, ..., \bm{X}_g]$, where $\bm{X} \in \mathbb{R}^{m \times n}$ and $n = \sum_{i=1}^{g} n_{i}$. In this case, the full results of PCA include $\bm{\Sigma} \in \mathbb{R}^{m \times n}$ and $\bm{V}^T = [\bm{V}_{1}^{T}, ... , \bm{V}_{g}^{T}] \in \mathbb{R}^{n \times n}$, where $\bm{V}_{i}^{T} \in \mathbb{R}^{n \times n_i}$ is the portion of the loadings matrix corresponding to the variables contributed by data holder $i$. 

The aim of this contribution is to design a privacy-preserving system that guarantees that (1) during the computation, data is not leaked to any other parties and (2) the loadings matrix $\bm{V}$ must be vertically and secretly distributed among data holders. While the first requirement is commonly shared by all privacy-preserving applications, the second requirement is more specific to federated MSPC. As explained in the previous chapter, loadings can reveal sensitive information about how variables interact with the principal components and can be used to calculate the contributions of variables to $T^2$ and $Q$ statistics. Therefore, data holder $i$ should know only $\bm{V}_{i} \in \mathbb{R}^{n_i \times n}$ that contains coefficients corresponding to its contributed variables. Furthermore, $\bm{V}_{i}$ has to be unknown to all other involved parties. In order to realize these goals, we propose an approach based on FedSVD \cite{chai2021}.

To apply FedSVD-based PCA as proposed in \cite{chai2021}, the system requires a Trusted Authority (TA) to handle key generation, and a Computation Service Provider (CSP) to take care of data aggregation and model building. Algorithm \ref{alg:fedpca_training} shows the overall model building workflow. Since $\bm{\Sigma}$, the number of NOC samples, the number of variables, and the number of principal components are shared by all data holders, the control limits for Hotelling's $T^2$ and $Q$-statistic can be estimated using Eq. \ref{eq:T2_calculation} and Eq. \ref{eq:Q_calculation} as for standard PCA.

After a FedPCA model is built, suppose a new sample $x =[x_1, ..., x_g]$, where $x_i \in \mathbb{R}^{1 \times n_i}$, is generated. Algorithm \ref{alg:fedpca_inference} is used to calculate scores, Hotelling's $T^2$ and $Q$-statistics, and the contribution of each variable to these indexes. Using these monitoring values, all data holders can check whether the sample is faulty or not, i.e. if the monitoring statistics lie above their critical limits. However, since $V_i$ is secretly owned by each data holder, the computation of Hotelling’s $T^2$- and $Q$- contributions can only be done locally. This might be greatly beneficial since all participating companies might know that a sample is faulty, and they can check whether the problem might be associated with their production line. However, each company only sees the contribution of its own variables to the fault, which provides feedback on how to improve its own process to benefit the entire value chain.

\begin{figure}[t!]
  \centering
  \includegraphics[width=0.9\textwidth]{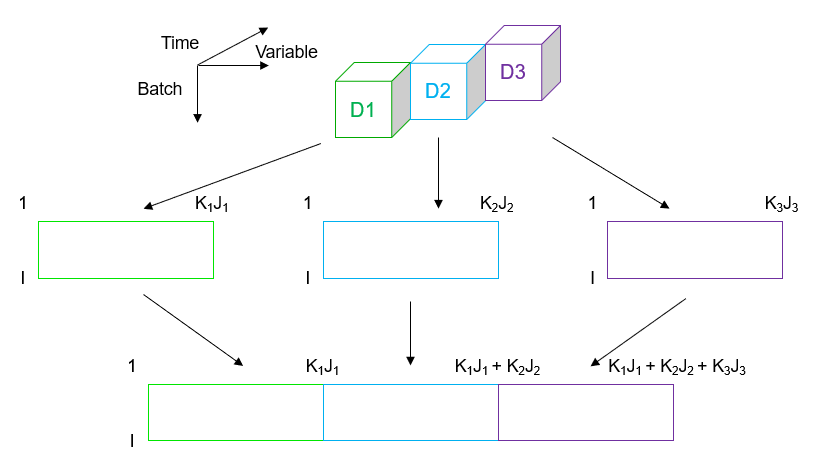}
  \caption{An illustration of how local data is mapped batch-wise in FedMPCA.}
  \label{fig:fedpca_unfolding}
\end{figure}

For batch process data, FedPCA cannot be applied directly. We thus propose an extension called FedMPCA that includes a data unfolding step before encryption and transfer to the CSP. However, unlike the transition from PCA to MPCA, not all unfolding techniques are applicable in the federated scenario. While variable-wise unfolding is invalid because local data do not share the same feature space, batch-wise unfolding is undertaken as shown in Figure \ref{fig:fedpca_unfolding}. 

Suppose there are $g$ participating data holders and data holder $i$ owns a batch data set of shape $I \times J_i \times K_i$ where $I$, $J_i$, and $K_i$ are the number of batches, the number of variables, and the number of time intervals respectively. When batch-wise unfolding is employed, the unfolded data of data holder $i$ is $\bm{X}_i \in \mathbb{R}^{m \times n_i}$ where $m = I$ and $n_i = K_iJ_i$. Therefore, the joined data $\bm{X} = [\bm{X}_1,\ldots, \bm{X}_g] \in \mathbb{R}^{m \times n}$, where $n = \sum_{i=1}^{g} n_{i} = \sum_{i=1}^{g} K_{i}J_{i}$. The full results of MPCA consist of $\Sigma \in \mathbb{R}^{m \times n}$ and $\bm{V}^T = [\bm{V}_{1}^{T},\ldots, \bm{V}_{g}^{T}] \in \mathbb{R}^{n \times n}$, where $\bm{V}_{i}^{T} \in \mathbb{R}^{n \times n_i}$ is the portion of the loadings matrix corresponding to the (unfolded) variables contributed by data holder $i$. Similar to FedPCA, at the end of FedMPCA model training, data holder $i$ receives $\bm{\Sigma}$ and $\bm{V}_{i}^{T}$ and can cooperate with the other data holders, the TA, and the CSP to calculate scores and the monitoring indexes.

In an online monitoring scenario, given all the preceding data holders have completed their processes, suppose the $i$-th data holder wants to calculate the monitoring indexes and variable contributions for an in-progress batch at time interval $k < K$. The corresponding data can be expressed as $x = [x_1, ..., \tilde{x}_i]$. The procedure is described in Algorithm \ref{alg:fedmpca_incomplete_batch}. Once the scores are obtained, data holders can calculate Hotelling's $T^2$, $Q$-statistics and contribution of variables to the two indexes in a similar manner as it is done for completed batches (described in lines 15-36 of Algorithm \ref{alg:fedpca_inference}).

\resizebox{0.99\textwidth}{!}{
\begin{algorithm}[H]
\DontPrintSemicolon
  
  \KwInput{$\bm{X} = [\bm{X}_1,\ldots, \bm{X}_g]$}
  \KwOutput{$\bm{\Sigma}$, $r$ and $\bm{V}^T = [\bm{V}_{1}^{T},\ldots, \bm{V}_{g}^{T}]$}
  \kwConstraint{Data holder $i$'s data is not leaked, and it receives $\bm{\Sigma}$, $r$ and $\bm{V}_{i}^{T}$ as results.}
  \Fn{\FFedPCATrain{$[\bm{X}_1,\ldots, \bm{X}_g]$}}{
         \KwTA(do:){
            Generate orthogonal matrices $\bm{P} \in \mathbb{R}^{m \times m}$, $\bm{B} \in \mathbb{R}^{n \times n}$. \\
            Then split $\bm{B}^T$ into $[\bm{B}_{1}^{T}, \ldots, \bm{B}_{g}^{T}]$ where $\bm{B}_i^T \in \mathbb{R}^{n \times n_{i}}$.
         }
         \KwDH(do:){
           \For{$i = 1 \rightarrow g$, Data holder $i$}{
                Download $\bm{P}$, $\bm{B}_i^T$ from TA and compute\\
                $\bm{X}_{i}^{'} = \bm{P}\bm{X}_{i}\bm{B}_{i}$
           }
         }
        \KwCSP(do:){
            Aggregate $\bm{X}^{'}$:\\
            $\bm{X}^{'} = \sum_{i=1}^{g} \bm{X}_{i}^{'} \quad \left(= \sum_{i=1}^{g} \bm{P}\bm{X}_{i}\bm{B}_{i} = \bm{PXB}\right)$\\
            Perform standard SVD:\\
            $\bm{X}^{'} = \bm{U}^{'} \bm{\Sigma} \bm{V}^{'T}$
         }
         \KwDH(do:){
           \For{$i = 1 \rightarrow g$, Data holder $i$}{
                Download $\bm{\Sigma}$ from CSP.\\
                Determine the number of principal components $r$ based on the cumulative sum of explained variance. \\
                Generate an random matrix $\bm{R}_{i} \in \mathbb{R}^{n_{i} \times n_{i}}$ \\
                Mask $\bm{B}_i^T$ through: $[\bm{B}^{T}_{i}]^{R} = \bm{B}^{T}_{i}\bm{R}_{i}$\\
                Send $[\bm{B}^{T}_{i}]^{R}$ to CSP.
           }
         }
         \KwCSP(wait to receive data and do:){
            \If{Receive $[\bm{B}^{T}_{i}]^{R}$ then}
            {
                Compute $[\bm{V}^{T}_{i}]^{R} = \bm{V}^{'T}[\bm{B}^{T}_{i}]^{R} \quad (= \bm{V}^{'T}\bm{B}^{T}_{i}\bm{R}_{i} = \bm{V}_{i}^{T}\bm{R}_{i})$ \\
                Send $[\bm{V}^{T}_{i}]^{R}$ back to data holder $i$.
            }
         }
         \KwDH(do:){
           \For{$i = 1 \rightarrow g$, Data holder $i$}{
                Receive $[\bm{V}^{T}_{i}]^{R}$ from CSP. \\
                Recover $\bm{V}_{i}^{T}$ by $\bm{V}_{i}^{T} = [\bm{V}^{T}_{i}]^{R}\bm{R}_{i}^{-1}$.
           }
         }
  }
\caption{FedPCA Training}
\label{alg:fedpca_training}
\end{algorithm}}

\newpage

\begin{algorithm}[H]
\DontPrintSemicolon
  
  \KwInput{$x = [x_1, ..., x_g]$}
  \KwOutput{$t_r$, $Q$-statistics, Hotelling's $T^2$, $Q_{cont} = [Q_{cont, 1}, ..., Q_{cont, g}]$, $T^{2}_{cont} = [T^{2}_{cont, 1}, ..., T^{2}_{cont, g}]$}
  \kwConstraint{Data holder $i$'s data is not leaked, and it receives $t_r$, $Q$-statistics, Hotelling's $T^2$, $Q_{cont, i}$, $T^{2}_{cont, i}$ as results.}
  \Fn{\FFedPCAPredict{$[x_1, ..., x_g]$}}{
         \KwTA(do:){
            Generate a random number $p$.
         }
         \KwDH(do:){
           \For{$i = 1 \rightarrow g$, Data holder $i$}{
                Download the random number $p$ from TA. \\
                Calculate local scores: $t_{r, i} = x_{i}\bm{V}_{r, i}$\\ 
                Encrypt the local scores with $p$: $t'_{r, i} = pt_{r, i} = px_{i}\bm{V}_{r, i}$
                Send $t^{'}_{r, i}$ to CSP.
           }
         }
        \KwCSP(do:){
            Aggregate $t^{'}_{r}$: $t^{'}_{r} = \sum_{i=1}^{g} t^{'}_{r, i} \quad \left(= p\sum_{i=1}^{g} x_{i}\bm{V}_{r, i} = px\bm{V}_{r}\right)$
         }
         \KwDH(do:){
           \For{$i = 1 \rightarrow g$, Data holder $i$}{
                Download $t^{'}_{r}$ from CSP.\\
                Recover the scores $t_{r}$: $t_{r} = \frac{t^{'}_{r}}{p}$\\
                Calculate $T^2$: $T^2 = t_{r}\bm{\Sigma}^{-2}t_{r}^{T}$ \\
                Calculate $T^{2}_{cont, i} = t_{r}\bm{\Sigma}^{-1}\bm{V}_{r, i}^{T}$\\
                Calculate local reconstruction errors $e_{i}$: $e_{i} = x_i - t_r\bm{V}_{r,i}^T$\\
                Calculate $Q_{cont, i}$: $Q_{cont, i} = e_{i}^2$\\
                Calculate local Q-statistics: $Q_{i} = e_{i}e_{i}^T$ \\
                Encrypt $Q_{i}$ using $p$: $Q^{'}_{i} = pQ_{i}$ \\
                Send $Q^{'}_{i}$ to CSP.
           }
         }
         \KwCSP(do:){
            Aggregate $Q^{'}$: $Q^{'} = \sum_{i=1}^{g} Q'_{i} \quad \left(= p\sum_{i=1}^{g} Q_{i} = pQ\right)$
         }
         \KwDH(do:){
           \For{$i = 1 \rightarrow g$, Data holder $i$}{
                Downloads $Q^{'}$ from CSP. \\
                Recover $Q$ by $Q = \frac{Q^{'}}{p}$
           }
         }
  }
\caption{FedPCA Inference}\label{alg:fedpca_inference}
\end{algorithm}

\newpage

\begin{algorithm}[H]
\DontPrintSemicolon
  
  \KwInput{$x = [x_1, ..., x_i]$}
  \KwOutput{$t_r^{[k]}$}
  \kwConstraint{Data holder $i$'s data is not leaked, and it receives $t_r^{[k]}$ as the result.}
  \Fn{\FFedMPCAPredict{$[x_1, ..., x_i]$}}{
         \KwTA(do:){
            Generate a random number $p$ and an random matrix $\bm{W} \in \mathbb{R}^{r \times r}$
         }
         \KwDH(do:){
           \For{$j = 1 \rightarrow g$, Data holder $j$}{
                Download the $p$ and $\bm{R}$ from TA. \\
                Calculate $t^{'}_{r, j}$ and $\bm{F}^{'}_{j}$ as follows:\\
                $t^{'}_{r, j} = px_{j}\tilde{\bm{V}}_{r, j}\bm{W}$\\
                $\bm{F}^{'}_{j} = \tilde{\bm{V}}_{r, j}^{T}\tilde{\bm{V}}_{r, j}\bm{W}$\\
                where $\tilde{\bm{V}}_{r, j} = \bm{V}_{r, j}$ for $j < i$, and $\tilde{\bm{V}}_{r, j} = \bm{V}_{r, j}[1:kJ_{j}]$ when $j = i$. Note that in this case, $\tilde{\bm{V}}_{r}^{T} = [\tilde{\bm{V}}_{r, 1}^{T}, ..., \tilde{\bm{V}}_{r, i}^{T}]$.\\
                Send $t^{'}_{r, j}$ and $\bm{F}^{'}_{j}$ to CSP.
           }
         }
        \KwCSP(do:){
            Calculates $t_r^{[k]'}$:
            \begin{align*}
                t_r^{[k]'} &= \sum_{j=1}^{i} px_{j}\tilde{\bm{V}}_{r, j}\bm{W} (\sum_{j=1}^{i} \tilde{\bm{V}}^{T}_{r, j}\tilde{\bm{V}}_{r, j}\bm{W})^{-1}\\
                           &\left(= p\sum_{j=1}^{i} x_{j}\tilde{\bm{V}}_{r, j}\bm{W}\bm{W}^{-1} (\sum_{j=1}^{i} \tilde{\bm{V}}^{T}_{r, j}\tilde{\bm{V}}_{r, j})^{-1}\right)\\
                           &\left(= px\tilde{\bm{V}}_{r}(\tilde{\bm{V}}^{T}_{r}\tilde{\bm{V}}_{r})^{-1}\right) \\
                           &\left(= pt_r^{[k]}\right)\\
            \end{align*}
            Broadcast $t_r^{[k]'}$ to all data holders.
         }
         \KwDH(do:){
           \For{$j = 1 \rightarrow g$, Data holder $j$}{
                Recover the real scores using $p$:\\
                $t_r^{[k]} = \frac{t_r^{[k]'}}{p}$
           }
         }
  }
\caption{FedMPCA Incomplete Batch}\label{alg:fedmpca_incomplete_batch}
\end{algorithm}

\section{Experiments}

As proof of concept, we applied FedPCA and FedMPCA to two industrial case studies from semiconductor manufacturing. The corresponding datasets SECOM\footnote{https://archive.ics.uci.edu/ml/datasets/SECOM} \cite{mccann2008} and ST-AWFD\footnote{https://github.com/STMicroelectronics/ST-AWFD} \cite{furnari2021} have been published previously and are in the public domain. 

\subsection{General settings} \label{general_settings}
For both case studies, we first divided the data ($\bm{X}$) variable-wise into two subsets ($\bm{X}_1$ and $\bm{X}_2$) corresponding to different process steps and assigned these to two (hypothetical) data holders. Subsequently, each subset was further split into a training, a validation and a test set, i.e. $\bm{X}_1 = \{\bm{X}_{1}^{train}, \bm{X}_{1}^{val}, \bm{X}_{1}^{test}\}$ and $\bm{X}_2 = \{\bm{X}_{2}^{train}, \bm{X}_{2}^{val}, \bm{X}_{2}^{test}\}$. The partition was done in a way that the training set only contained NOC batches and the validation and test sets consisted of both NOC and faulty batches. The training set was used for training the model. The validation set was used to set control limits for Hotelling’s $T^2$ and $Q$-statistic. By means of a grid search, we selected the lowest thresholds with the highest F1 score on the validation set as the control limits. The test set was utilized to evaluate the model performance. Four models were built to simulate three common real-world situations:
\begin{itemize}
    \item Situation 1: One company has access to all data $\bm{X}$ and can use that data to build a fault detection model based on PCA (MPCA).
    \item Situation 2: Each company only has access to private data and can use that data to build a local fault detection model to detect faults that occurred in their production line. In this case, Company 1 owns the PCA1 (MPCA1), and Company 2 owns the PCA2 (MPCA2) model. If a fault is detected by one of the models, it is considered to be detected.
    \item Situation 3: Each company only has access to private data, however, the two companies cooperate to build a federated fault detection model based on FedPCA (FedMPCA).
\end{itemize}

Table \ref{table:general_settings_data_split} shows the training and test set used for each model. The number of principal components was chosen such that the cumulative sum of explained variance was 90\%. 

\begin{table}[ht!]
\centering
\caption{Training set and test set for each model.}
\begin{tabular}{| >{\centering}p{3.5cm} | >{\centering}p{2.5cm} | >{\centering}p{2.5cm} | >{\centering\arraybackslash}p{2.5cm} |} 
 \hline
 Model & Training Data & Validation Data & Test Data \\ [0.5ex] 
 \hline
 PCA1 (MPCA1) & $\bm{X}_{1}^{train}$ & $\bm{X}_{1}^{val}$ & $\bm{X}_{1}^{test}$ \\ 
 \hline
 PCA2 (MPCA2) & $\bm{X}_{2}^{train}$ & $\bm{X}_{2}^{val}$ & $\bm{X}_{2}^{test}$ \\ 
 \hline
 PCA (MPCA) & $\{\bm{X}_{1}^{train}, \bm{X}_{2}^{train}\}$ & $\{\bm{X}_{1}^{val}, \bm{X}_{2}^{val}\}$ & $\{\bm{X}_{1}^{test}, \bm{X}_{2}^{test}\}$ \\ 
 \hline
 FedPCA (FedMPCA) & $\bm{X}_{1}^{train}$, $\bm{X}_{2}^{train}$ & $\bm{X}_{1}^{val}$, $\bm{X}_{2}^{val}$ & $\bm{X}_{1}^{test}$, $\bm{X}_{2}^{test}$ \\ 
 \hline
\end{tabular}
\label{table:general_settings_data_split}
\end{table}

We benchmarked the models in terms of effectiveness to detect faulty batches and also fault diagnoses on the test set. A batch was considered faulty when either the $T^2$ or the $Q$-statistic exceeded the predefined control limits. The effectiveness was evaluated by the F1 score and the fault diagnosis was evaluated based on Hotelling’s $T^2$- and $Q$-contribution plots.

\subsection{Case study 1: SECOM Dataset}
\subsubsection{Data description.}
SECOM is a static dataset consisting of 1567 observations, each with 590 variables ($S_1$ to $S_{590}$) and one label for the quality test (-1 means the observation is normal and 1 indicates the observation is faulty). As with any real-world dataset, SECOM contains missing values and irrelevant variables that have to be addressed before modeling. As the focus in this case study is fault detection and diagnosis, for data cleaning, we considered the 38 variables recommended by \cite{arif2013}, and dropped all instances that contain missing values. According to \cite{arif2013}, based on the property of the semiconductor manufacturing monitoring process, the selected variables can be divided into five workstations. In this experiment, we assumed that $\bm{X}_1$ and $\bm{X}_2$ include all parameters of the first three and the last two workstations, respectively. Table \ref{table:secom_variable_split} and Table \ref{table:secom_train_test_split} show a summary of each dataset.

\begin{table}[ht!]
\centering
\caption{List of variables that belong to each data holder.}
\begin{tabular}{| >{\centering}p{5em} | >{\centering}p{2.5cm}| >{\centering\arraybackslash}p{7.5cm} |} 
 \hline
 Dataset & No. Variables & Variable Name \\ [0.5ex] 
 \hline
 $\bm{X}_1$ & 21 & $S_{15}$, $S_{27}$, $S_{33}$, $S_{36}$, $S_{48}$, $S_{60}$, $S_{62}$, $S_{64}$, $S_{118}$, $S_{122}$, $S_{124}$, $S_{125}$, $S_{131}$, $S_{134}$, $S_{145}$, $S_{153}$, $S_{184}$, $S_{201}$, $S_{206}$, $S_{288}$, $S_{342}$ \\ 
 \hline
 $\bm{X}_2$ & 17 & $S_{421}$, $S_{426}$, $S_{427}$, $S_{430}$, $S_{435}$, $S_{454}$, $S_{461}$, $S_{470}$, $S_{478}$, $S_{492}$, $S_{511}$, $S_{520}$, $S_{525}$, $S_{560}$, $S_{569}$, $S_{572}$, $S_{574}$\\
 \hline
\end{tabular}
\label{table:secom_variable_split}
\end{table}

In this experiment, the optional validation set was not used where the control limits for $T^2$ and $Q$-statistic are set using Eq. \ref{eq:T2_calculation} and \ref{eq:Q_calculation}.

\begin{table}[ht!]
\centering
\caption{Summary of training and test set.}
\begin{tabular}{| >{\centering}p{2.5cm} | >{\centering}p{3cm} | >{\centering}p{3cm} | >{\centering\arraybackslash}p{3cm} |} 
 \hline
 Dataset & No. NOC samples & No. faulty samples & No. features \\ [0.5ex] 
 \hline
 $\bm{X}_{1}^{train}$ & 488 & 0 & 21 \\ 
 \hline
 $\bm{X}_{1}^{test}$ & 74 & 48 & 21 \\
 \hline
  $\bm{X}_{2}^{train}$ & 488 & 0 & 17 \\
 \hline
  $\bm{X}_{2}^{test}$ & 74 & 48 & 17 \\
 \hline
\end{tabular}
\label{table:secom_train_test_split}
\end{table}

\subsubsection{Evaluation.}

The performance of all evaluated models is shown in Table \ref{table:secom_model_performance}. PCA and FedPCA show the same performance. This is expected because, in \cite{chai2021}, the authors prove that FedSVD is a lossless method that produces the same results as standard SVD. Notably, FedPCA outperforms PCA1, PCA2, and their combination by achieving a higher F1 score which underpins the benefit of integrated vs. local process models.

\begin{table}[ht!]
\centering
\caption{Model performance on SECOM data.}
\begin{tabular}{| >{\centering}p{2.5cm} | >{\centering}p{1cm} | >{\centering}p{1cm} | >{\centering}p{1cm} | >{\centering}p{1cm} | >{\centering\arraybackslash}p{1.5cm} |}
 \hline
 Model & TP & TN & FP & FN & F1 score\\ [0.5ex] 
 \hline
 PCA & 19 & 55 & 19 & 29 & 0.44 \\ 
 \hline
 FedPCA & 19 & 55 & 19 & 29 & 0.44 \\
 \hline
 PCA1 + PCA2 & 14 & 53 & 21 & 34 & 0.33 \\ 
 \hline
\end{tabular}
\label{table:secom_model_performance}
\end{table}

\begin{figure}
  \centering
  \includegraphics[width=1.0\textwidth]{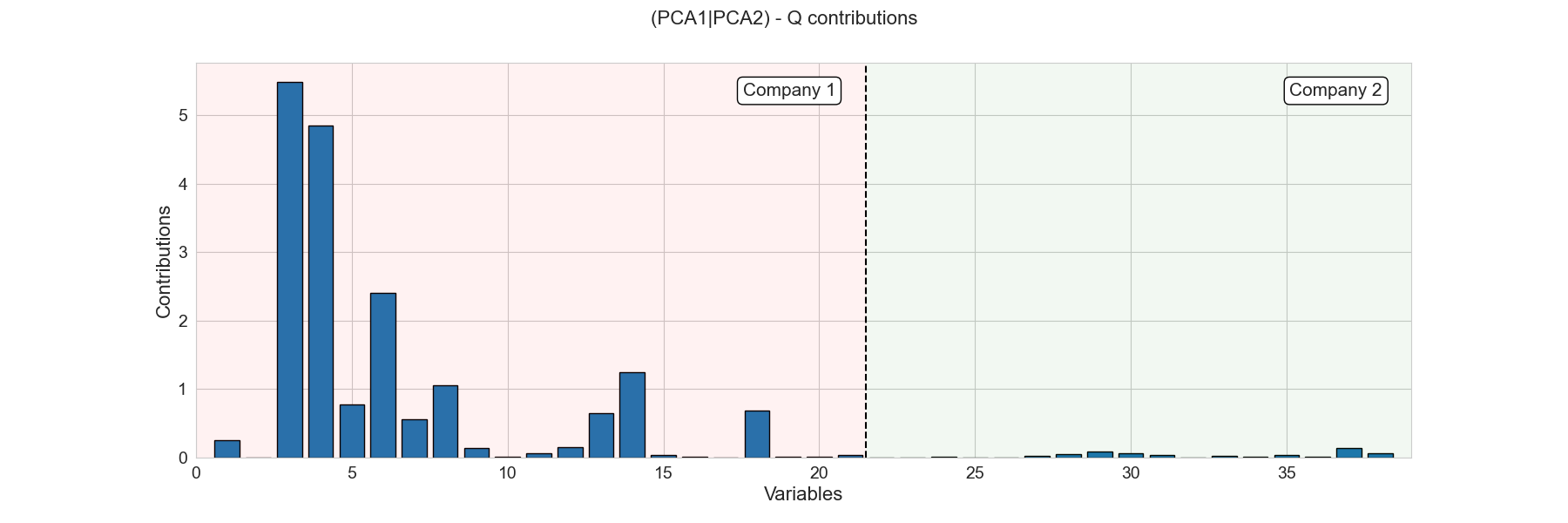}
  \includegraphics[width=1.0\textwidth]{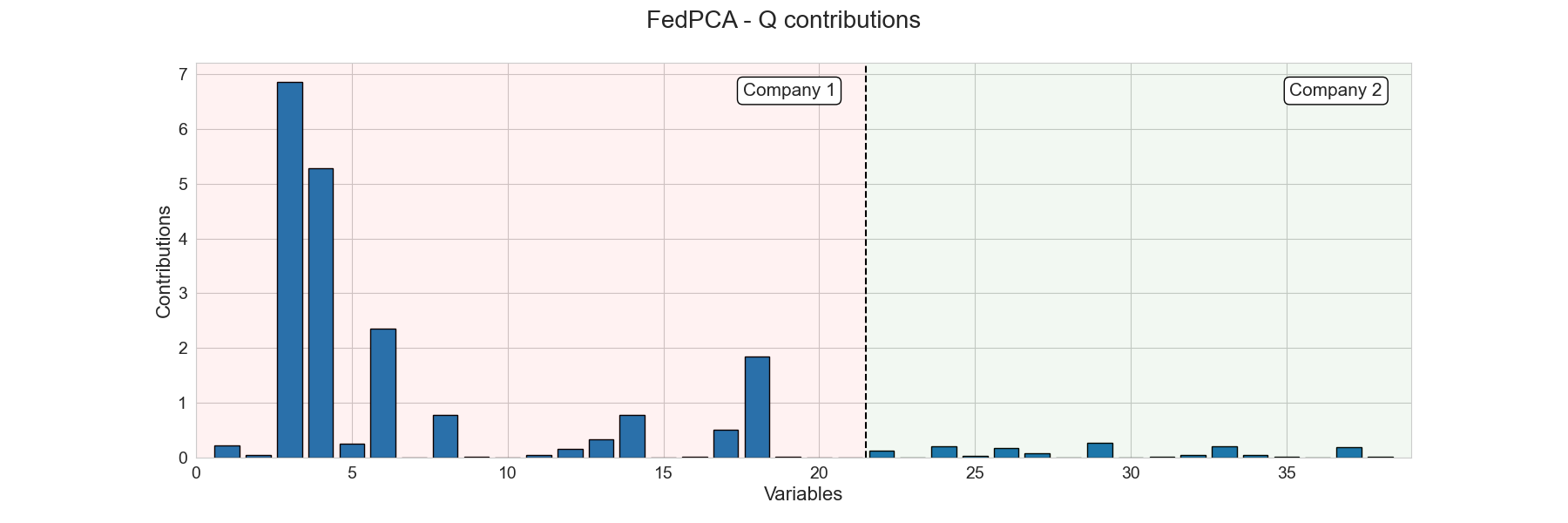}     
  \caption{Q contribution plots generated for Sample 1.}
  \label{fig:secom_contribution_plot_sample_1}
\end{figure}

\begin{figure}
  \centering
  \includegraphics[width=1.0\textwidth]{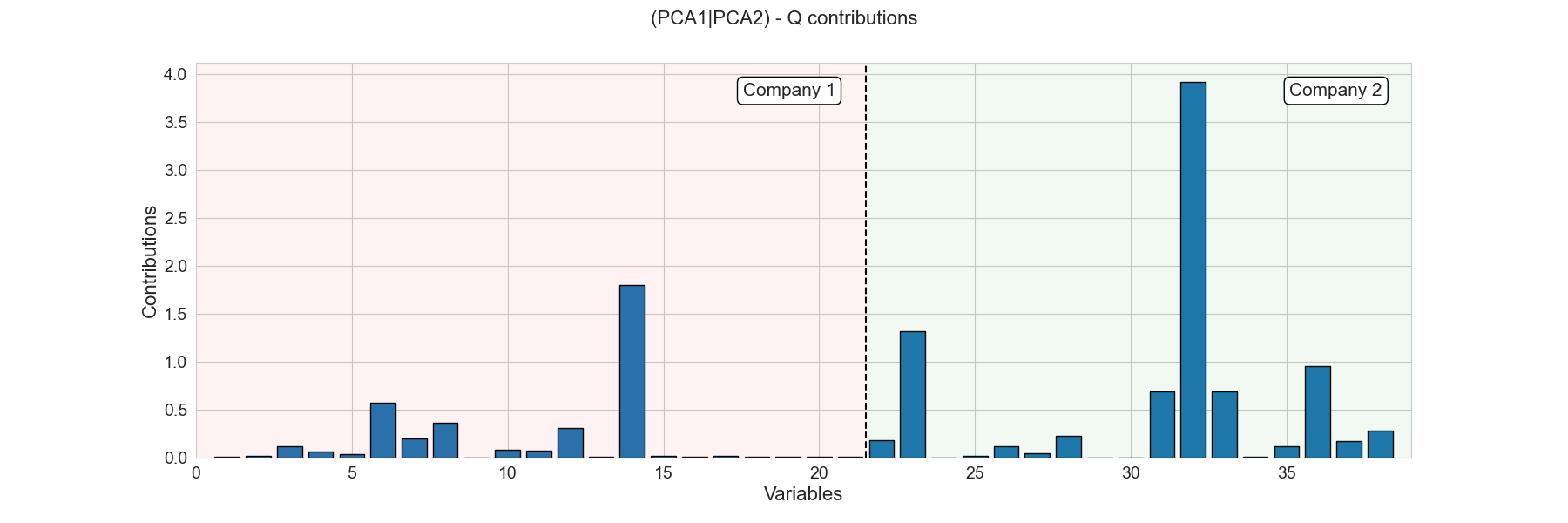}
  \includegraphics[width=1.0\textwidth]{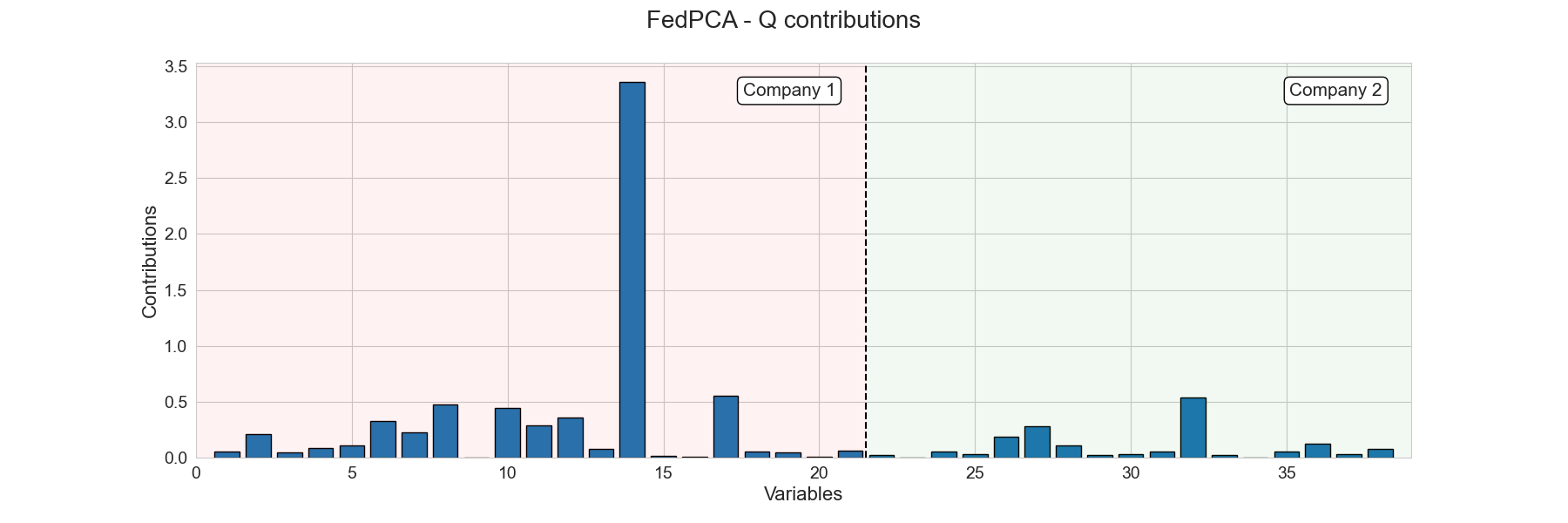}     
  \caption{Q contribution plots generated for Sample 2.}
  \label{fig:secom_contribution_plot_sample_2}
\end{figure}

\begin{figure}
  \centering
  \includegraphics[width=1.0\textwidth]{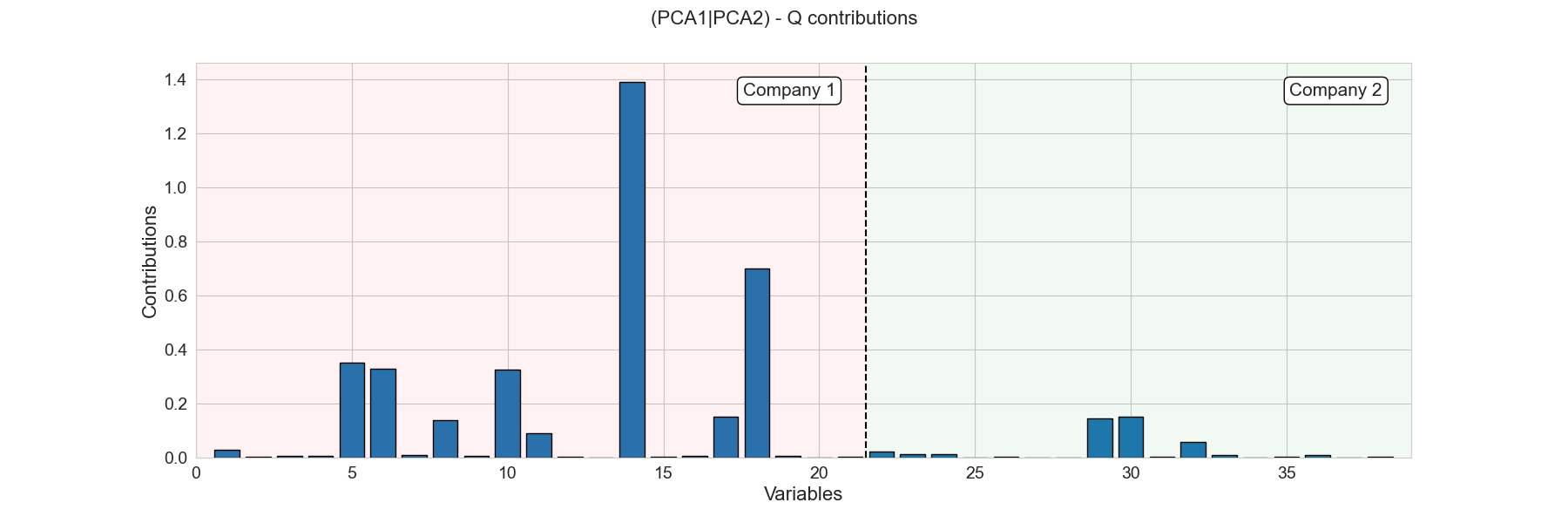}
  \includegraphics[width=1.0\textwidth]{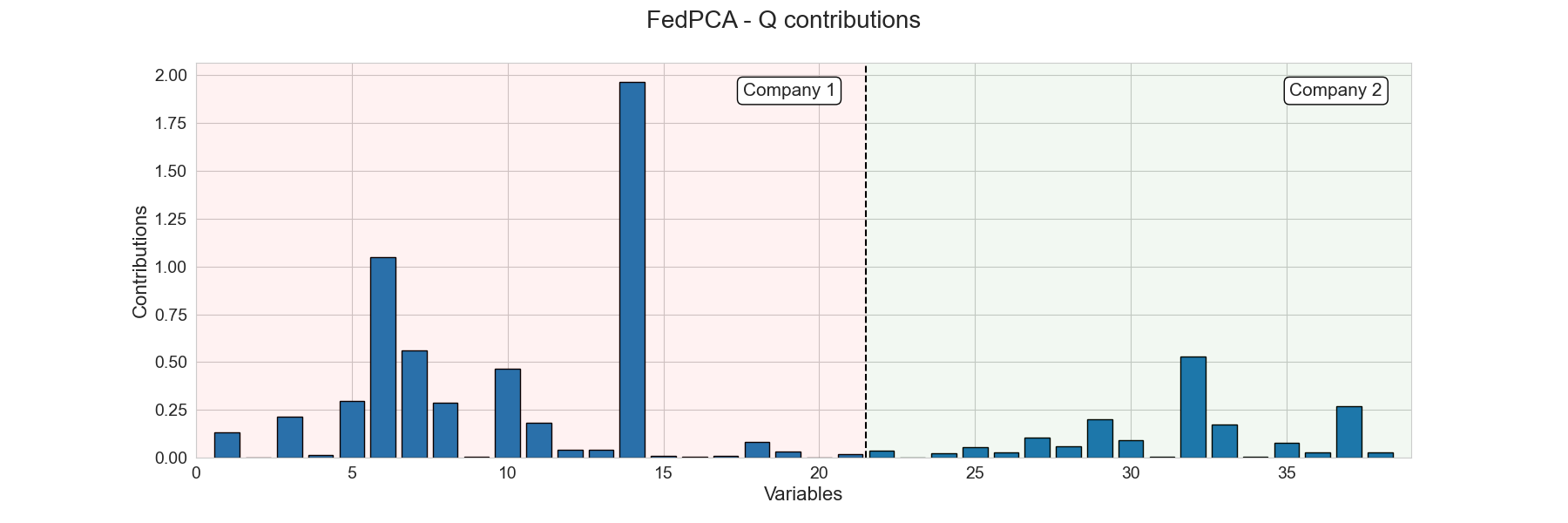}     
  \caption{Q contribution plots generated for Sample 3.}
  \label{fig:secom_contribution_plot_sample_3}
\end{figure}

To evaluate the capability of our approach to diagnose faults, we used the trained models to generate contribution plots for Hotelling's $T^2$ and $Q$-statistic. Figures \ref{fig:secom_contribution_plot_sample_1}, \ref{fig:secom_contribution_plot_sample_2}, and \ref{fig:secom_contribution_plot_sample_3} show contribution plots of some selected faults. In each of these figures, there are two plots. The upper plot shows the contribution of each input variable calculated by PCA1 and PCA2 (i.e. the local models). The lower one shows the variable contributions calculated by FedPCA. In the middle of each plot, there is a vertical dashed line representing the (hypothetical) company border. Note that each company can only reconstruct the contributions corresponding to the variables that they own using their private data ($x_i$), private loadings matrix ($\bm{V}_i$), and the shared matrix $\bm{\Sigma}$.

Figures \ref{fig:secom_contribution_plot_sample_1} and \ref{fig:secom_contribution_plot_sample_2} show two examples of faulty products that were detected by both FedPCA and the combination PCA1 + PCA2. In Figure \ref{fig:secom_contribution_plot_sample_1}, it can be seen that for Sample 1, the two plots are quite similar and variables of Company 1 show a much higher contribution to the fault than those of Company 2 indicating that the fault is caused predominantly by the latter. Variable \#3 and \#4 are reasonable candidates for further root cause analysis. In contrast, in Figure \ref{fig:secom_contribution_plot_sample_2} FedPCA and the local models disagree in terms of the variables that contribute to the fault. Whereas the local models suggest a significant contribution from both parties (with variables from company 2 in fact showing higher overall contributions), the FedPCA model indicates that the fault is mostly associated with company 1. While Variable \#14 shows the highest impact according to PCA1 and FedPCA, the most contributed variable according to PCA2 is Variable \#32.

Figure \ref{fig:secom_contribution_plot_sample_3} demonstrates an example where FedPCA detected a faulty sample that passed both PCA1 and PCA2. Even though the $Q$-statistic calculated by PCA1 is high, it doesn't surpass the control limit. For PCA2, it is clear that the $Q$-statistic is small. This example represents cases where the problem is caused by not only the process parameters of one data holder but by a combination/interaction of process parameters across the (hypothetical) company border. An advantage of FedPCA, in this case, is that it can make use of all available data to increase performance, and at the same time, the root cause (sensitive information) is known by only the data holder, and completely unknown by other parties. While the second company can reliably claim the problem did not come from their process production and request the first company to do a checkup. It won't know the exact contribution of input variables of the first company. On the other hand, the first company can use the contribution plots together with their know-how to optimize their machine settings to reduce out-of-specification events later in the value chain.

\subsection{Case study 2: ST-AWFD Dataset}

\subsubsection{Data description.}
ST-AWFD is a batch dataset that contains a total of 1156 batches with 20 variables and an average of 100 samples per batch. Each batch is labeled as normal or faulty through a temporal reference window. The production process is divided into two steps called Step 1 and Step 2. Depending on the batch, the length of each step might differ. 

In order to apply batch-wise MPCA and FedMPCA, all the batches must have the same length. Therefore, in this experiment, we only selected batches with lengths of 110 consisting of 65 observations for Step 1 and observations for Step 2. After the data cleaning phase, there are 966 batches left which include 648 NOC batches and 318 faulty batches. A summary of the data partition is shown in Table \ref{table:st_train_test_split}.

\begin{table}[ht!]
\centering
\caption{A summary of training, validation, and test set used in the experiment.}
\begin{tabular}{| >{\centering}p{2cm} | >{\centering}p{2cm} | >{\centering}p{2cm} | >{\centering}p{2cm} | >{\centering\arraybackslash}p{2cm} |} 
 \hline
 Dataset & No. NOC samples & No. faulty samples & No. features & No. time intervals \\ [0.5ex] 
 \hline
 $\bm{X}_{1}^{train}$ & 482 & 0 & 20 & 65 \\
 \hline
  $\bm{X}_{1}^{val}$ & 83 & 159 & 20 & 65 \\
 \hline
 $\bm{X}_{1}^{test}$ & 83 & 159 & 20 & 65 \\
 \hline
  $\bm{X}_{2}^{train}$ & 482 & 0 & 20 & 45 \\
  \hline
  $\bm{X}_{2}^{val}$ & 83 & 159 & 20 & 45 \\
 \hline
  $\bm{X}_{2}^{test}$ & 83 & 159 & 20 & 45 \\
 \hline
\end{tabular}
\label{table:st_train_test_split}
\end{table}

The control limit for the $T^2$ statistic was calculated by Eq. \ref{eq:T2_calculation}. For the $Q$-statistic, initially, Eq. \ref{eq:Q_calculation} was used for determining the confidence limit. However, we found that the returned threshold was over-optimistic and led to a high number of false positives for all models. Therefore, the control limit for the $Q$-statistic was obtained by cross-validation using the training and validation set (described in \ref{general_settings}) instead.

\subsubsection{Evaluation}

The performance of all evaluated models are shown in Table \ref{table:st_model_performance}. Similar to case study 1, MPCA and FedPCA returned the same performance and outperformed the combination of MPCA1 and MPCA2. In this experiment, the contribution plots were also generated for faulty batches. However, due to the large number of variables, it is difficult to judge the difference between plots (results not shown).

\begin{table}[ht!]
\centering
\caption{Model performance on ST-AWFD data.}
\begin{tabular}{| >{\centering}p{3cm} | >{\centering}p{1cm} | >{\centering}p{1cm} | >{\centering}p{1cm} | >{\centering}p{1cm} | >{\centering\arraybackslash}p{1.5cm} |}
 \hline
 Model & TP & TN & FP & FN & F1 score\\ [0.5ex] 
 \hline
 MPCA & 159 & 83 & 0 & 0 & 1 \\ 
 \hline
 FedMPCA & 159 & 83 & 0 & 0 & 1 \\
 \hline
 MPCA1 + MPCA2 & 159 & 74 & 9 & 0 & 0.97 \\ 
 \hline
\end{tabular}
\label{table:st_model_performance}
\end{table}
Altogether, our results on the two case studies underpin the benefit of federated, PCA-based process modeling in terms of better fault detection performance and more informative fault diagnosis that takes into account the interactions between process parameters across (hypothetical) company borders.

\section{Conclusion}

In the present work, we proposed a framework towards enabling privacy-preserving, federated multivariate statistical process control (FedMSPC) of process chains involving multiple consecutive process steps operated by different companies. In particular, we have employed federated PCA following the batch-wise unfolding of the horizontally concatenated (encrypted) datasets from the participating parties and demonstrated the application of such models for federated fault detection and privacy-preserving fault diagnosis. To the best of our knowledge, this is the first study that proposes a solution to this problem.

\section*{Acknowledgements}
We would like to thank Dr. Mohit Kumar for fruitful discussions on privacy preserving machine learning and to Dr. Marco Reis for support regarding dataset selection.
The research reported in this paper has been partly funded by the Federal Ministry for Climate Action, Environment, Energy, Mobility, Innovation and Technology (BMK), the Federal Ministry for Digital and Economic Affairs (BMDW), and the State of Upper Austria in the frame of SCCH, a center in the COMET - Competence Centers for Excellent Technologies Program managed by the Austrian Research Promotion Agency FFG and
the FFG project circPlast-mr (Grant No. 889843).

\end{document}